\pdfoutput=1
\documentclass[10pt]{article}
\newif\ifarxiv
\arxivtrue
\ifarxiv \usepackage[preprint]{tmlr} \else \usepackage{tmlr} \fi
\ifarxiv
  \newcommand{\availabilitytext}{The code, run configurations, and the data behind every figure and table (CSV, and the rendered example images) are on GitHub at \url{https://github.com/nilotpal09/improving-signals-declining-encoder}; the encoder checkpoints are on Zenodo at \url{https://doi.org/10.5281/zenodo.23666713}. Training is not bitwise reproducible, so the checkpoints are part of the release.}
  \newcommand{\llmusagetext}{Large language models assisted with coding, experiment execution, analysis,
figure production, and manuscript drafting and editing under the author's
direction and review. Study design, verification of the reported results, and
scientific interpretation remained with the author, who takes responsibility
for the work.}
\else
  \newcommand{\availabilitytext}{The code, run configurations, and the data behind every figure and table (CSV, and the rendered example images) are in the supplementary material. The encoder checkpoints are publicly archived; we leave out the link to keep the review anonymous. Training is not bitwise reproducible, so the checkpoints are part of the release.}
  \newcommand{\llmusagetext}{Large language models assisted with coding, experiment execution, analysis,
figure production, and manuscript drafting and editing under the authors'
direction and review. Study design, verification of the reported results, and
scientific interpretation remained with the authors, who take responsibility
for the work.}
\fi
\usepackage{graphicx,booktabs,amsmath,amssymb,xcolor,longtable}
\usepackage[colorlinks=true, linkcolor=blue, citecolor=blue, urlcolor=blue]{hyperref}
\newcommand{\cosfloorRounded}{0.996} %
\newcommand{\untrainedThirty}{0.407} %
\newcommand{\untrainedHundred}{0.559} %
\newcommand{\untrainedThreeHundred}{0.691} %
\newcommand{\untrainedThousand}{0.837} %
\newcommand{\roomThreeHundred}{0.30} %
\newcommand{\roomThousand}{0.16} %
\newcommand{\nThirtyDrops}{0.22--0.26} %
\newcommand{\nHundredDrops}{0.14--0.28} %
\newcommand{\nThirtyHighPeaks}{0.91--0.93} %
\newcommand{\nThirtyLowPeaks}{0.70, 0.83 and 0.71} %
\newcommand{\wbarvec}{(0.513,\,-0.645,\,0.567)} %
\newcommand{\ndial}{\ensuremath{\mu}}   %

\newcommand{\eventMainHorizon}{300} %
\newcommand{\eventSweepSize}{56} %
\newcommand{\eventSweepAligned}{55} %
\newcommand{\eventLeadingCCA}{0.98} %
\newcommand{\eventRankOneRange}{0.955--0.994} %
\newcommand{\eventAnatomySize}{13} %
\newcommand{\eventLearnedBelow}{12 of 13} %
\newcommand{\eventRankCentred}{11 of 13} %
\newcommand{\eventRankPR}{11 of 13} %
\newcommand{\eventProjectorDim}{16} %
\newcommand{\eventNullWindows}{101} %
\newcommand{\eventNullRuns}{43} %
\newcommand{\eventNullMax}{0.107} %
\newcommand{\eventShallowDepth}{0.302} %
\newcommand{\eventTestSize}{2{,}000} %
\newcommand{\eventTestDifference}{0.02} %

\newcommand{\fitEvents}{4{,}000} %
\newcommand{\evalEvents}{2{,}000} %
\newcommand{\anchorGap}{20} %
\newcommand{\auditVariants}{21} %
\newcommand{\auditMatched}{26} %
\newcommand{\auditHeldout}{13} %

\newcommand{\selectionLossRounded}{0.21} %
\newcommand{\selectionAlphaFlat}{0.029} %
\newcommand{\selectionFinalFlat}{0.0006} %
\newcommand{\readPeakMedian}{0.97} %
\newcommand{\readFinalMedian}{0.77} %
\newcommand{\costZone}{0.15} %
\newcommand{\costZtwo}{0.29} %
\newcommand{\costE}{0.20} %
\newcommand{\readDeclineCut}{0.07} %
\newcommand{\respDecline}{25} %
\newcommand{\respHealthy}{10} %
\newcommand{\respDeclineMuZero}{19} %
\newcommand{\respHealthyMuZero}{4} %
\newcommand{\auditBestTimely}{9} %
\newcommand{\auditCalibOnsets}{7} %

\newcommand{\setupGrid}{64} %
\newcommand{\setupElements}{100} %
\newcommand{\setupNanoParams}{0.7M} %
\newcommand{\setupWidthSmall}{384} %
\newcommand{\setupWidthLarge}{2048} %
\newcommand{\setupEarlyStride}{2} %
\newcommand{\setupDenseUntil}{20} %
\newcommand{\setupLateStride}{10} %
\newcommand{\setupEpochExamples}{64{,}000} %
\newcommand{\setupEpochBatches}{500} %
\newcommand{\setupBatchSize}{128} %
\newcommand{\setupRidgeAlpha}{1} %

\newcommand{\mechanismSNRInitCos}{0.86} %
\newcommand{\mechanismSNRPreCos}{0.91} %

\newcommand{\methodPatch}{4} %
\newcommand{\methodTokenSide}{16} %
\newcommand{\methodNanoDim}{96} %
\newcommand{\methodNanoDepth}{6} %
\newcommand{\methodHeads}{3} %
\newcommand{\methodMLPRatio}{4} %
\newcommand{\methodTinyDim}{192} %
\newcommand{\methodTinyDepth}{12} %
\newcommand{\methodQueryParams}{85k} %

\newcommand{\lidarShareRounded}{98\%} %
\newcommand{\lidarCorrRounded}{0.98} %
\newcommand{\lidarSharedRise}{24} %
\newcommand{\lidarSharedLambdaFall}{7} %
\newcommand{\lidarNoiseEnergy}{12 to 14 of 21} %

\newcommand{\methodChannels}{4} %
\newcommand{\methodInputScale}{5500} %
\newcommand{\methodEnergyLow}{20{,}000} %
\newcommand{\methodEnergyHigh}{40{,}000} %
\newcommand{\methodMAEMask}{0.75} %
\newcommand{\methodMAEDepth}{4} %
\newcommand{\methodMAEDim}{48} %
\newcommand{\methodRandomDirections}{2{,}000} %
\newcommand{\methodOracleFeatures}{24} %

\ifarxiv
  \newcommand{\compcite}[1]{\citep[#1]{kakati2026shared}}
  \hypersetup{pdfauthor={Nilotpal Kakati},
              pdftitle={Improving Signals, Declining Encoder: A Known-Answer Test of Label-Free Monitoring in Self-Supervised Learning}}
\else
  \newcommand{\compcite}[1]{\citep[#1]{kakati2026shared}}
  \hypersetup{pdfauthor={}, pdftitle={Under review at TMLR}}
\fi

\title{Improving Signals, Declining Encoder: A Known-Answer Test of Label-Free Monitoring in Self-Supervised Learning}
\ifarxiv
\author{\name Nilotpal Kakati \email nilotpal.kakati09@gmail.com \\
        \addr Weizmann Institute of Science, Rehovot, Israel}
\else
\author{}
\fi
\begin{document}
\maketitle

\begin{abstract}
Self-supervised training is monitored without labels, through the loss
and statistics of the learned features. We report a case where we know
the encoders lost part of what they had learned, so we can check whether
these signals noticed. We trained LeJEPA encoders on synthetic images
with known generating factors. In nine of ten seeds, the readability of
those factors fell abruptly and stayed down, nearly always along the same
direction in factor space. Meanwhile the loss kept falling, and most of
the eleven label-free signals we tracked kept improving and rated the
declining runs better than runs that did not decline. Standard change-point
detection rarely flagged the decline near its onset, and selecting
checkpoints by these signals mostly chose ones after it.
\end{abstract}

\section{Introduction}\label{sec:intro}

Self-supervised training has no labels to check against, so it is judged
by its loss and by label-free statistics of the features. SimSiam
\citep{chen2020simsiam} uses feature standard deviation to diagnose
collapse. RankMe \citep{garrido2023rankme} and LiDAR
\citep{thilak2024lidar} are used to select models, and LiDAR's authors
present it as better than RankMe at spotting the best point in a
training run.
\citet{arputharaj2026metrics} compare many such metrics
against downstream accuracy across trained models. None of these
statistics measures whether an encoder keeps what it has already learned.
Nor can that be checked on natural data, because nobody knows what an
encoder should encode at each checkpoint.

\begin{figure}[t]\centering
\includegraphics[width=0.93\linewidth]{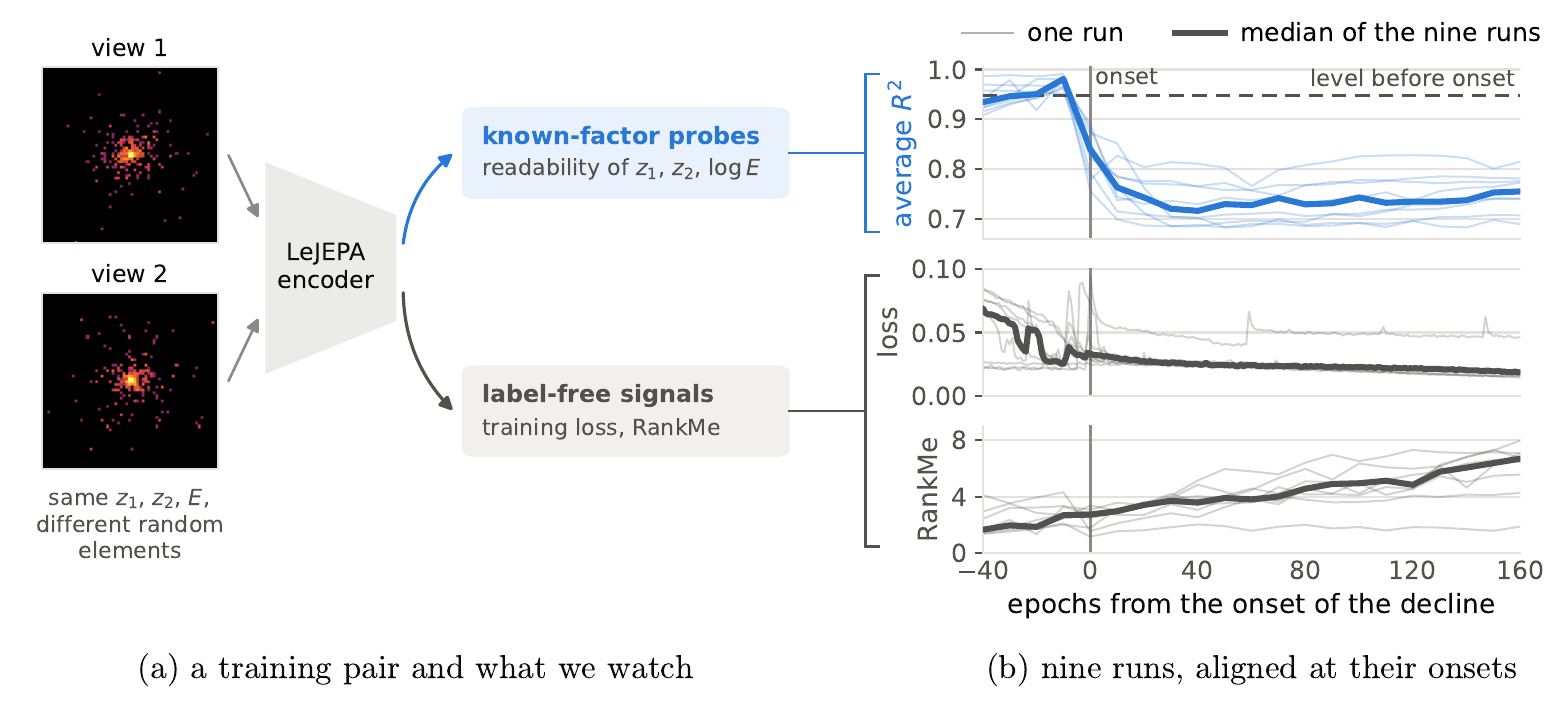}
\caption{(a) One training pair at $N=100$: two renders with the same shape
factors $z_1,z_2$ and energy $E$ that differ only in the sampled elements
(log intensity, summed over channels). Grey: label-free signals; blue:
readability, the average held-out ridge $R^2$ of the three factors. (b) The
nine default runs that declined, aligned at their onsets, which are
estimated from readability. Readability drops at onset and stays below its
level before onset, while the training loss keeps its downward trend, with
spikes in some runs, and RankMe rises. Figure~\ref{fig:eventloss} shows all
ten runs against training epoch.}
\label{fig:overview}\label{fig:schematic}
\end{figure}

An image generator with known factors removes that problem: the
readability of each factor can be measured from frozen features at every
checkpoint. On one such generator, we trained encoders with a reduced
two-view LeJEPA \citep{lejepa2025} recipe and saw them learn the factors
and then abruptly lose part of that readability (Figure~\ref{fig:overview}). Over
the same epochs the training loss kept falling and RankMe, a measure of
effective rank, rose. Falling effective rank is the sign of dimensional
collapse \citep{jing2022dimensional,garrido2023rankme}, so this decline did
not look like a collapse,
and a check for collapse would have read it as progress. The two training views share the factors, so
the objective never asked the encoder to discard them. The decline
recurred across seeds and held up under nonlinear probes and a separately
generated test split. A masked autoencoder \citep[MAE;][]{he2022mae} trained on the same images for the
same number of epochs did not decline.

Since we know when the encoder lost readability, we can check whether
label-free signals noticed. We tracked eleven of them around each decline
and compared the runs that declined with runs of the same settings that
did not. Most kept improving through the decline and rated the runs that
declined better than those that did not (Section~\ref{sec:event}). Even the
best pairing of signal and change-point method, chosen after the fact,
alarmed near the onset in only about a third of the declines; most pairings
alarmed in far fewer.
Choosing checkpoints by these signals mostly picked ones after the
decline, with worse readability. A LiDAR variant we computed by mistake
flagged most declines, but it relies on information a real training run does
not have, so it cannot serve as a monitor (Section~\ref{sec:lidar}).

\section{Experimental Setup and Measurements}\label{sec:rig}

\paragraph{Images and views.} We use a simpler configuration of the
simulator of \citet{kakati2026shared}, built on PyCaloFlash v1.0
\citep{dreyer_pycaloflash}, to generate sparse images from two shape
coordinates, $(z_1,z_2)$, and energy $E$. Each image has \methodChannels{} channels of
$\setupGrid\times\setupGrid$ pixels and shows one pattern at a fixed
position. The shape coordinates set how the pattern's intensity is spread
over the channels: where it peaks and how broad the peak is. $E$ sets the
overall intensity and shifts the typical peak position and width
(Appendix~\ref{app:generator}). In each channel, $N$ points, called
elements, are drawn at random from this intensity profile and binned into
the pixels. The element count controls image graininess (Figure~\ref{fig:rig}). Each image's shape
coordinates are drawn from a distribution set by a parameter $\mu$. At
$\mu=0$ each coordinate is standard normal; as $\mu$ grows, its distribution
splits into two Gaussians whose centres move apart at fixed overall variance,
and near $\mu=1$ they barely overlap.
Instead of image augmentations, the
two training views are independent renders with the same shape and energy;
only the sampled elements differ.
Figure~\ref{fig:overview}a shows a training pair and the two measurement
paths, and Appendix~\ref{app:generator} describes the generator's physical
model.

\paragraph{Training configuration.} We trained a small ViT
\citep{dosovitskiy2021image} encoder,
ViT-Nano, with a projector using the LeJEPA objective
\citep{lejepa2025}, which combines agreement between the projected training
views with SIGReg, a regularizer that pushes projector outputs toward a
standard normal distribution. Our reduced recipe uses two whole-image views. The default projector has
hidden width \setupWidthSmall{}; a wide projector, of width
\setupWidthLarge{}, is the second main configuration. Unless stated
otherwise, runs last \eventMainHorizon{} epochs and use $N=\setupElements$
and $\mu=0$, and default runs also use the default projector. Every epoch
draws fresh images from the generator. MAE \citep{he2022mae} on the same
generator provides a comparison with another training objective, and
ViT-Tiny \citep{touvron2021training} provides an architecture check. Appendix~\ref{app:training} gives
the full training details.

\paragraph{Measuring readability.} At each saved checkpoint, we freeze the
encoder and use its final-layer tokens, layer-normalized and averaged over
positions, as features. Readability is the average held-out $R^2$ of ridge
readouts of $z_1$, $z_2$ and log energy from these features. The readouts are fitted on one set of generated
images and scored on a separate set, both matched to training in $\mu$ and
$N$. These probe images are single renders, not pairs.
The same images are reused across checkpoints and across runs of the same
condition. Effective rank is measured with RankMe
\citep{garrido2023rankme}, the exponential of the entropy of the normalized
singular values of the same features. Appendix~\ref{app:training} gives the
readout implementation, and Appendix~\ref{app:metricdetails} the metric
conventions.

\paragraph{Decline and onset.} Drop from peak is the largest
readability from epoch 20 on minus the final value, and a run declines if
this drop is at least \readDeclineCut{}. The onset is the first checkpoint
after the peak at which readability has fallen by a fifth of the drop. We fixed
this rule after inspecting the trajectories of most of our runs, and applied
it unchanged to runs trained later (Appendix~\ref{app:training}). The pre-onset checkpoint is the last one at
least \anchorGap{} epochs before onset.

\section{Readability Declines While Label-Free Signals Improve}\label{sec:event}

We first show the decline, then use it to test the label-free signals:
how they behaved around it (Section~\ref{sec:declinesignals}), whether
change-point detection flagged it in time, and which checkpoints they would
have chosen (Section~\ref{sec:alarms}).

\subsection{The Decline and the Signals Around It}\label{sec:declinesignals}

\begin{figure}[!htbp]\centering
\includegraphics[width=\linewidth]{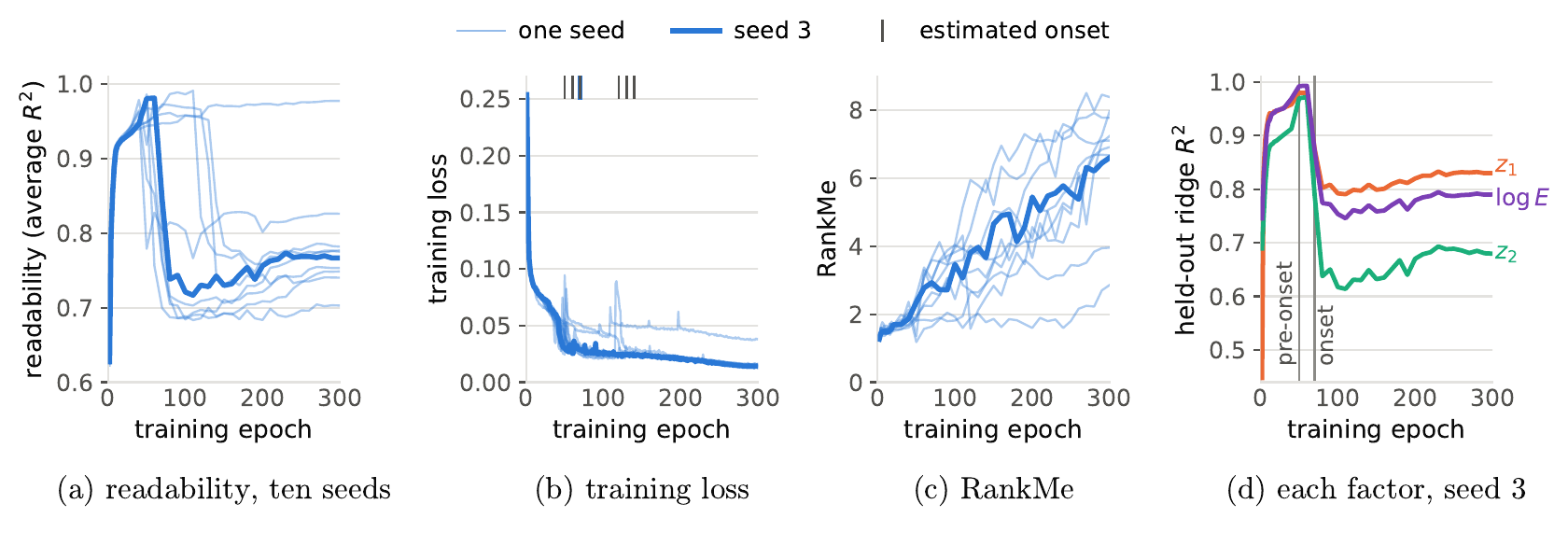}
\caption{The ten default runs over 300 epochs: (a) readability, (b) training
loss and (c) RankMe. RankMe increases overall, with local decreases. It
stays far below the feature dimension of 96, partly because it does not
subtract the large feature mean. (d) Held-out ridge $R^2$ of each factor for seed 3, the median
decline in both drop and onset, highlighted in the other panels; its
pre-onset checkpoint comes before the peak.}
\label{fig:eventloss}\label{fig:anatomy}
\end{figure}

Readability rose and then fell sharply in nine of ten default runs
(Figure~\ref{fig:eventloss}a\label{loc:eventfig}). The decline also occurred
with the wide projector (Figure~\ref{fig:comparisons}). Every decline lowered the readability of all
three factors (Figure~\ref{fig:anatomy}d shows one run), and nearly every decline lost the same combination of them
(Section~\ref{sec:direction}). Onsets varied across seeds; some runs
partly recovered, but none came back to its peak. Over the same epochs
the training loss kept its downward trend and RankMe rose
(Figure~\ref{fig:eventloss}b, c).

\begin{figure}[!t]\centering
\includegraphics[width=\linewidth]{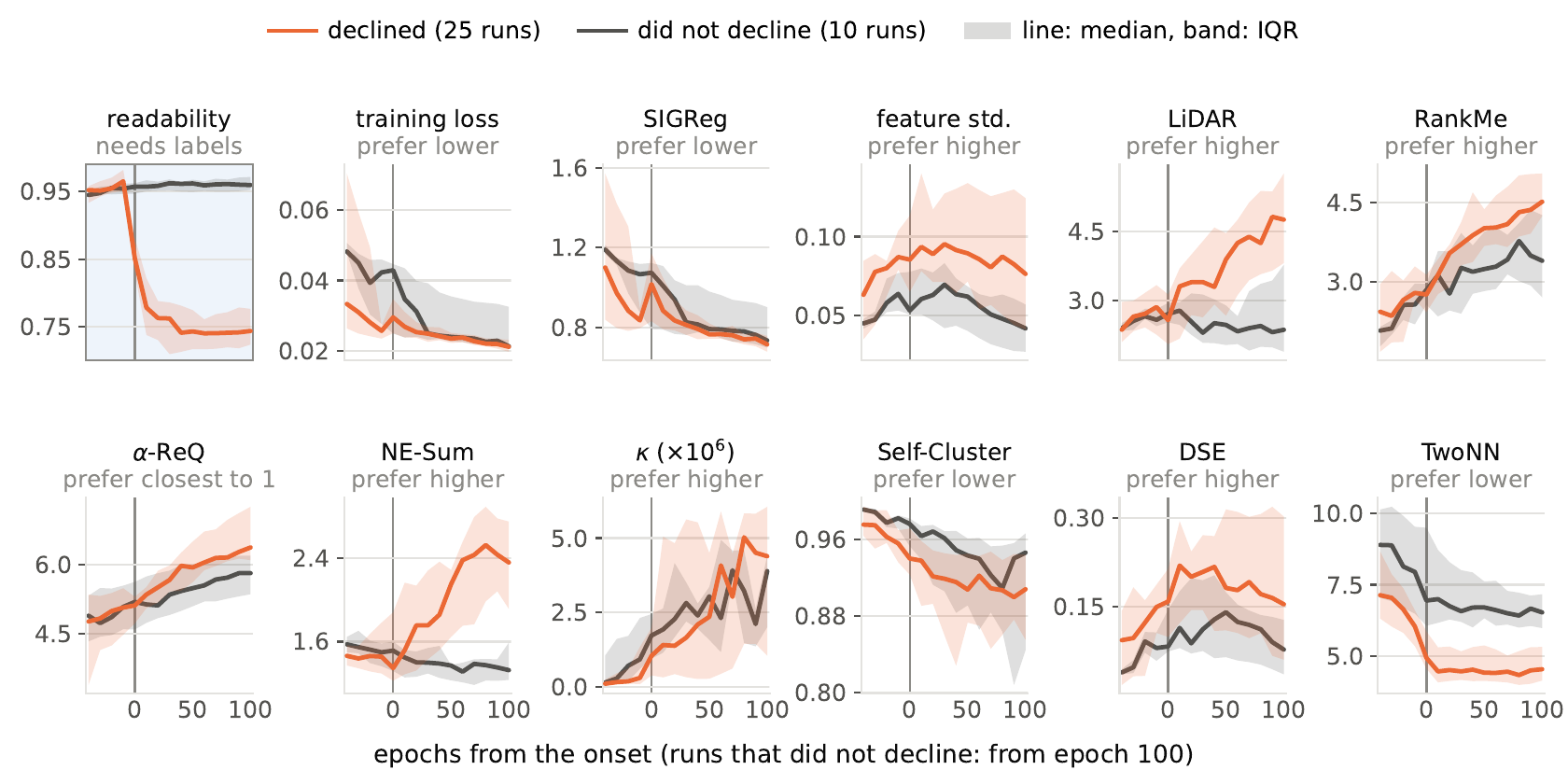}
\caption{Label-free signals around a known readability decline. Orange:
decline runs from the settings of the alarm audit
(Appendix~\ref{app:chronological}) that have both outcomes, aligned at their
onsets. Grey: runs without a decline from the same settings, aligned at
epoch 100, close to the median onset. Readability falls and stays down, while most
label-free signals keep moving in their preferred direction; $\alpha$-ReQ
moves away from 1. RankMe and the bottom row are the seven metrics
compared by \citet{arputharaj2026metrics}.}
\label{fig:response}
\end{figure}

We follow the signals on the runs we use to test alarms
(Appendix~\ref{app:chronological}). Figure~\ref{fig:response} compares the
runs that declined with those that did not, within the settings (projector
and $\mu$) that had both outcomes. We align the decline runs at their onsets and the runs
without a decline at epoch 100, close to the median onset.
The figure follows eleven label-free signals around the decline: the training loss,
SIGReg, feature standard deviation, LiDAR \citep{thilak2024lidar} and the
seven metrics compared by \citet{arputharaj2026metrics}, which summarize
the spectrum of the features, how features relate across images, or their
intrinsic dimension. LiDAR is an effective rank computed over groups of
inputs (Section~\ref{sec:lidar}). Each panel of Figure~\ref{fig:response}
gives its signal's preferred direction, and Table~\ref{tab:metricconventions}
lists the metrics' sources and variants.

After the onset, most signals kept moving in their preferred direction,
and eight of the eleven rated the decline runs better than the runs without
a decline at every checkpoint after the onset. For RankMe, the two medians
nearly tied at the first checkpoint after the onset. The exceptions were
$\alpha$-ReQ, which rated the decline runs worse throughout (further from
1), and the loss and $\kappa$, which were mixed. So by most of these
signals, the runs that lost readability looked like the better runs.

The runs without a decline have no onset. Aligning them anywhere from epoch
50 to 160 instead leaves most of these verdicts unchanged, but not SIGReg's
(Appendix~\ref{app:metricdetails}). The two groups also differ in mix, with
more $\mu=0$ runs among the decline runs.

\subsection{Alarms and Checkpoint Choice}\label{sec:alarms}

We ran three standard change-point detection methods (CUSUM,
\citealp{page1954continuous}; EWMA, \citealp{roberts1959control};
Page--Hinkley, \citealp{hinkley1971inference}) on every signal, with
thresholds set on control runs from
another shape distribution (Appendix~\ref{app:chronological}). Even the
best combinations of signal and method, picked after the fact, alarmed
within ten epochs of onset in only about a third of the declines
(\auditBestTimely{} of \auditMatched{} scored), and they also alarmed in some
runs without a decline.

Without a timely alarm, these signals can still be used to pick a
checkpoint.\label{sec:selection} We asked which saved checkpoint the loss, LiDAR and each metric would
select: the one where it was best in its preferred direction. The
cost of a choice is the readability it loses relative to the best
candidate, chosen with hindsight. This is an offline comparison, not a
stopping rule.

Signals that keep improving through training pick late checkpoints. Loss
minimization and RankMe picked a checkpoint at or after the onset in every
decline run, and LiDAR in all but one. Each of the three cost a
median $\selectionLossRounded{}$ in readability, the same as taking the
final checkpoint.
LeJEPA proposes its loss as evidence of
representation quality \citep{lejepa2025}; here the loss improved while
readability declined. $\alpha$-ReQ picks early, which costs little in
decline runs but loses readability in runs without a decline, where the
final checkpoint loses almost none (Appendix~\ref{app:metricdetails}). The
other five metrics chose checkpoints at or after onset in most
decline runs, in every variant we computed (Table~\ref{tab:metricconventions}),
at about the same cost as the final checkpoint. So, apart from $\alpha$-ReQ,
choosing a checkpoint by these signals did about as well as simply taking
the last one.

\section{Checking and Understanding the Decline}\label{sec:checks}

We first check that the decline does not depend on the readout and that the
rise in effective rank does not depend on how it is measured
(Section~\ref{sec:readoutchecks}), then show that nearly all declines share
one direction (Section~\ref{sec:direction}), and then report which changes
to training altered the decline and which did not (Section~\ref{sec:where}).

\subsection{Readout and Objective Checks}\label{sec:readoutchecks}

These checks use \eventAnatomySize{} decline runs, nine with the default
projector and four with the wide one. The drop shows under every readout we
tried: pooled ridge, an attention probe, an MLP, and token-level
ridges. The two nonlinear probes did not beat pooled ridge in most
comparisons, and rescoring on a separately generated test split changed
little. A readout
we did not try could still recover the lost readability
(Appendices~\ref{app:training} and~\ref{app:eventchecks}).

The rise in effective rank does not depend on how it is measured.
RankMe rose from the pre-onset to the final checkpoint in all
\eventAnatomySize{} runs, and in \eventRankCentred{} after subtracting the
feature mean. RankMe on the projector output stayed near its maximum at both
checkpoints, so the projector did not lose rank either
(Appendix~\ref{app:eventchecks}). A masked autoencoder trained on the same
images for as long did not decline; all ten runs ended at their peak.

\subsection{A Shared Direction}\label{sec:direction}

Each run's direction is the combination of factors whose readout $R^2$
fell most between the pre-onset and final checkpoints
(Appendix~\ref{app:direction}).
Across all our runs, every
decline but one lost readability along almost the same direction in factor
space. The exception declined twice, and
splitting it after seeing the trajectory puts its first decline along the
common direction and its second nearly perpendicular to it. Runs without a decline
lost far less along their own direction: $R^2$ fell by at most
\eventNullMax{}, against at least \eventShallowDepth{} in every decline
(Appendix~\ref{app:eventchecks}).

The common direction mixes all three factors with comparable weights,
which is why every factor's readout falls (Figure~\ref{fig:anatomy}d).
Other combinations stayed readable. Canonical correlation analysis
between the features and the factors finds three combinations; in the runs
of Section~\ref{sec:readoutchecks}, the first two stayed almost perfectly
correlated while only the third fell. In each of these runs, the change in
readout error lay almost entirely along one direction
(Appendix~\ref{app:eventchecks}). So each encoder lost one combination of the
factors and kept the others readable.

\subsection{Changes That Altered the Decline}\label{sec:where}

\begin{figure}[!t]\centering
\includegraphics[width=0.66\linewidth]{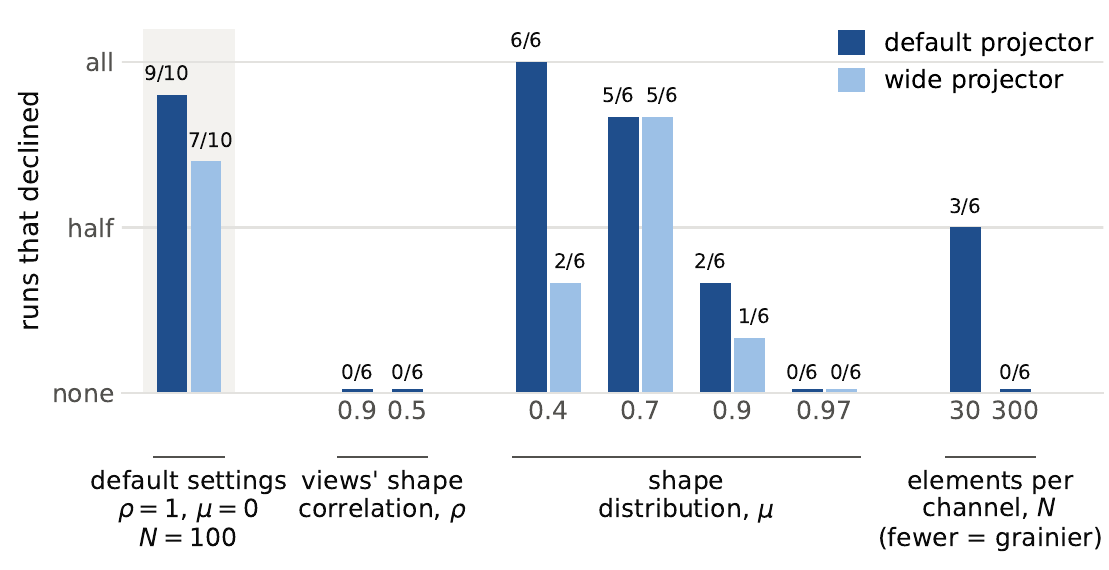}
\caption{How many runs declined under each change, out of how many runs
(the count on each bar). The left group uses the default settings, and each
other group changes one of them. A run declined if its readability ended at
least \readDeclineCut{} below its peak. All runs trained for 300 epochs.
Appendix~\ref{app:occurrence} covers the conditions not shown.}
\label{fig:comparisons}\label{fig:viewrefresh}\label{fig:occurrence}\label{fig:hardness}
\end{figure}

Partially refreshing shape between the two training views removed the
decline (Figure~\ref{fig:comparisons}\label{loc:comparisonsfig}). In these runs
the two views' shape coordinates are only partly correlated instead of
equal, while energy stays shared (Appendix~\ref{app:occurrence}). No refreshed run declined, against nine of
ten default runs on the same images and probes, and refreshed runs trained
for 1000 epochs also stayed near their peaks.

The shape distribution mattered too (Figure~\ref{fig:comparisons}). The
decline was common at $\ndial=0.4$ and $0.7$ and uncommon within 300 epochs
at $\ndial=0.9$, but every $\ndial=0.9$ run trained for 1000 epochs declined.
No run at $\ndial=0.97$ declined, even after 1000 epochs, though some fell
well below their peak mid-run and then recovered.

Image graininess mattered as well (Figure~\ref{fig:comparisons}). No run
declined with less grainy images, although these runs peaked far enough
above what an untrained encoder already reads for a typical drop to show.
With the cleanest images we tried, runs did not peak that far above it, so
they say nothing about the decline. With grainier images, half the runs
declined about as much as at the default graininess, and the rest learned
poorly from the start (Appendix~\ref{app:occurrence}).

Changes to the optimizer, the objective and the encoder did not remove the
decline. It occurred without weight decay and under a constant learning
rate, though with three seeds each we cannot say whether either change makes
it more or less common. With the SIGReg weight raised to the value
recommended by \citet{lejepa2025}, three of ten runs declined, and in those
the loss kept falling and RankMe rose. One of two runs with a ViT-Tiny
encoder declined, along the same direction. Table~\ref{tab:occurrence-summary}
lists these runs.

\section{LiDAR Flags the Decline When Its Groups Redraw Shape}\label{sec:lidar}

LiDAR measures how evenly the differences between groups of inputs
spread across feature directions, after discounting the variation
inside each group. Normally, each group holds views made the same way as
the training views. By mistake, our first evaluation built its groups with an
augmentation from the paper that introduced the simulator \compcite{\S2}:
every render in a group keeps the group's energy but
gets a newly drawn shape, whereas training views share both. With groups
that redraw shape, LiDAR flagged most declines on time, with few false
alarms; with training-view groups, it rarely did
(Figure~\ref{fig:auditcomparison}). With shape redrawn inside each
group, the groups differ mainly in energy, so the largest direction in
LiDAR's spectrum follows energy. As the decline lowers
energy readability, that direction weakens, the spectrum evens out, and
LiDAR rises (Appendix~\ref{app:lidarchecks}).

This augmentation is not part of standard LiDAR. Building it requires
knowing which factors are shape and which is energy, and changing one
while holding the other fixed. And because LiDAR with these groups rises
during the decline, it reads the decline as an improvement: used as a
quality score, as in Section~\ref{sec:selection}, it also preferred
checkpoints after the decline.

\section{Related Work}\label{sec:related}

The label-free signals we track were mostly validated by comparing models.
RankMe \citep{garrido2023rankme} and LiDAR \citep{thilak2024lidar} were
tested on trained models and sets of checkpoints, and LeJEPA's training
loss correlated with downstream linear-probe accuracy over training
configurations \citep{lejepa2025}. Within a single run of I-JEPA
\citep{assran2023ijepa}, \citet{thilak2024lidar} also found that effective rank peaked early and then
fell, with the best downstream accuracy far from that peak. The closest
work to ours, \citet{arputharaj2026metrics}, compares label-free metrics
with downstream accuracy for many trained models and finds that each
metric's reliability depends on architecture and training objective. We
ask a different question, whether these signals notice when a single run
loses what it learned, so our result does not contradict theirs. Online
linear probes, as in solo-learn \citep{dacosta2022sololearn}, follow what
the features encode during training, but they need labels, as do the
probes that measure the decline here.

Known generating factors have been used before to test what unsupervised
models learn. \citet{locatello2019challenging} trained thousands of models
on factor datasets and found that well-disentangled models seemingly could
not be identified without supervision; our checkpoint-choice result is the
within-run counterpart. For LeJEPA, \citet{klindt2026world} prove that the
objective's optimum recovers Gaussian generating factors when the two
views' factors are related but not identical. Our shared views are
identical, where the objective no longer favours recovering the factors,
and the theorem concerns the optimum rather than the path training takes,
so it says nothing either way about our declines or our refresh runs. The main guarantee of the original LeJEPA paper
\citep{lejepa2025}, that isotropic Gaussian embeddings serve downstream
probes best, concerns how the embeddings are distributed, not which factors
they keep.

\citet{simon2023stepwise} find that some joint-embedding methods learn their
embeddings in steps, one or a few directions at a time. Our declines also happen along a single direction,
here in factor space, but remove readability instead of adding it.

\section{Discussion}\label{sec:mechanism}

Effective rank measures how evenly the features spread over directions,
not what they encode, so it can rise while readability falls. RankMe
does not subtract the feature mean, and in these runs the mean was the
largest direction of the features' spectrum; its share fell, which made
the spectrum more even. RankMe
recomputed after subtracting the mean also rose in most runs
(Section~\ref{sec:readoutchecks}), so the mean does not explain the whole
rise.

Refreshing shape between the two views removed the decline
(Section~\ref{sec:where}), even though energy stayed shared and the common
direction includes energy. So the decline is not about shape alone. One
possible cause is feature suppression, where easy features hold back the
learning of others \citep{chen2021intriguing,robinson2021shortcut}. Here
it would have to remove a feature that had already been learned. Two
observations fit this. In an untrained encoder, and in one run before its
onset, the common direction was close to the direction the two views agreed
on least. Ridge readouts from the untrained encoder, and from an oracle
that uses simple image summaries such as the total intensity in each
channel, also found it harder to read than most directions. So the common
direction was weak from the start, as suppression would need. But the
hardest direction for both puts little weight on energy, so neither
observation explains the energy in the common direction
(Appendices~\ref{app:eventchecks} and~\ref{app:occurrence}).

Optimizer instabilities are also possible
\citep{lewkowycz2020catapult,thilak2022slingshot,wortsman2023smallscale}:
Section~\ref{sec:where} shows only that weight decay and a changing
learning rate are not necessary for the decline.

\section{Limitations}\label{sec:limitations}

The evidence comes from one image generator, small encoders and a reduced
two-view LeJEPA recipe, and MAE is the only other objective that learned
on it. It does
not show how common the decline is on natural images or with the
recommended recipe at scale \citep{lejepa2025}. A contrastive
model trained with SimCLR's loss \citep{chen2020simclr} never moved away
from its initialization on this generator, and changing the learning rate
did not help, so we do not know whether contrastive training shows the
decline. Readability is what our probes recover, so the decline shows that
the factors became harder to read; it does not show that the information
is gone. We also tested only three
standard change-point detection methods, with thresholds set on control runs from
another shape distribution. That they rarely alarmed in time does not
show that the decline cannot be detected from these signals.

\section{Conclusion}\label{sec:conclusion}

In this LeJEPA case study, the encoders lost readability they had
learned, while the training loss kept falling and most label-free signals
kept improving. Change-point detection on these signals caught a third of the declines in
time at best, and checkpoints chosen by loss or RankMe came
at or after the onset, with worse readability than earlier ones. A signal
that keeps improving therefore does not show that an encoder keeps what it
has learned.

\section*{Code and Data Availability}
\availabilitytext

\section*{LLM Usage}
\llmusagetext

\ifarxiv
\section*{Acknowledgements}
The author thanks Eilam Gross for discussions and feedback on the manuscript.
This work was supported by the Minerva Grant No. 7015027, the Knell Family
Institute for Artificial Intelligence, the Benoziyo Center for High Energy
Physics, a BSF Grant No. 2024220, and the Krenter-Perinot Center for High
Energy Particle Physics.
\fi

\bibliography{references}
\bibliographystyle{tmlr}

\clearpage
\appendix
\raggedbottom %
\section{The Generator}\label{app:generator}
We use a single-particle configuration of the simulator of
\citet[\S2 and App.~A]{kakati2026shared}, whose images show several
overlapping particles. The simulator uses
PyCaloFlash v1.0 \citep{dreyer_pycaloflash}, a simplified model of a particle shower in a calorimeter, a layered detector
that absorbs a high-energy particle and records where its energy is
deposited. Each of the \methodChannels{} channels of the
$\setupGrid\times\setupGrid$ image is one detector layer, at increasing
depth, and each pixel holds the energy deposited in one cell of that layer.
Each layer samples $N$
elements (individual deposits) independently from a smooth intensity
profile, and the two views of a training pair share the profile and differ
only in these samples. Energies are in MeV (megaelectronvolts). Raw cell
values $x$ are transformed as $\log(1+x)/\log(1+\methodInputScale)$.

The particle's energy $E$ is drawn with density proportional to $E^{-2}$ on
$[\methodEnergyLow,\methodEnergyHigh]$ MeV. The deposit's spread over depth
follows a gamma-shaped curve \citep{grindhammer1993parameterized} with two
parameters: $T$, the depth at which it
peaks, and $\alpha$, which sets how wide the peak is. $E$ scales every
deposit, so it sets the overall intensity. It also sets the means, widths and correlation of the map from
the shape coordinates to $\log T$ and $\log\alpha$.

\begin{figure}[!htbp]\centering
\includegraphics[width=\linewidth]{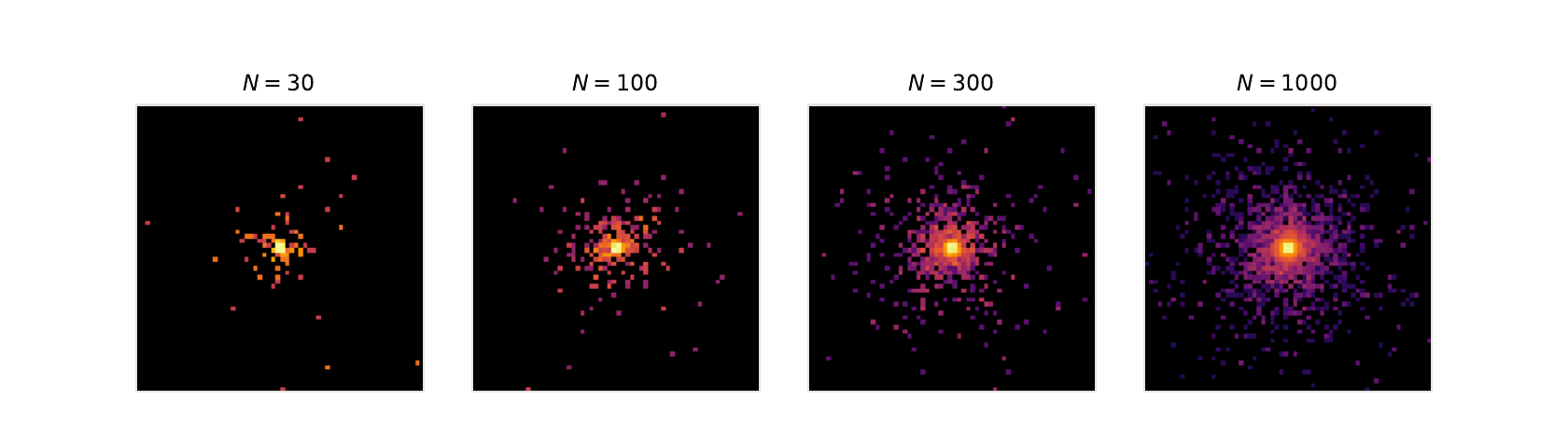}
\caption{One image rendered at $N=30$, 100, 300 and 1000, with the
intensity summed over the \methodChannels{} depth layers and shown as
$\log(1+\text{intensity})$ on one shared scale. The $N=100$ render is view~1 of
Figure~\ref{fig:overview}a.}
\label{fig:rig}
\end{figure}

Each shape coordinate is drawn independently as
\[
z=s\,\ndial+\sqrt{1-\ndial^2}\,g,
\]
where $s=\pm1$ with equal probability and $g$ is standard normal. At
$\ndial=0$ this gives a standard normal, and increasing $\ndial$ separates
the two mixture components while keeping zero mean and unit variance.
Figure~\ref{fig:rig} shows one image at the four element counts used in the
paper. %

\section{Training and Probe Details}\label{app:training}
ViT-Nano (\setupNanoParams{} parameters) uses patch size \methodPatch{}, a $\methodTokenSide\times\methodTokenSide$
token grid, embedding dimension \methodNanoDim{}, \methodNanoDepth{} blocks,
\methodHeads{} attention heads and MLP ratio \methodMLPRatio{}.
ViT-Tiny uses dimension \methodTinyDim{} and \methodTinyDepth{} blocks
(5.40M parameters). We used AdamW \citep{loshchilov2019decoupled} with batch
size \setupBatchSize{} in
bfloat16. Each epoch has \setupEpochExamples{} freshly generated examples in
\setupEpochBatches{} batches, each example supplying two views, and
checkpoints are saved every \setupEarlyStride{} epochs to epoch
\setupDenseUntil{} and every \setupLateStride{} after. The
learning rate warms up from $2\times10^{-4}$ to $10^{-3}$ over 40 epochs
and then decays to $10^{-6}$ on a cosine schedule, and weight decay rises
from 0.04 to 0.4 on a cosine schedule. The projector has two
Linear--BatchNorm--ReLU hidden layers of width 384 (default) or 2048 (wide), followed by a
linear map to 16 outputs. BatchNorm statistics are pooled over both views
and all images in a batch. Default-projector runs of 300 epochs took three to
seven GPU-hours on NVIDIA RTX A5000 or A6000 GPUs.

\paragraph{Reduced training recipe.} The objective is
$\lambda\,\mathrm{SIGReg}+(1-\lambda)\,\mathrm{agreement}$,
where agreement is the mean squared distance of each view's projection
from the mean over views. Compared with the starting points LeJEPA
recommends \citep{lejepa2025}, we used two views rather than ten,
$\lambda=0.02$ rather than $0.05$, 256 SIGReg
projections rather than 1024, and the integration range $[-3,3]$ rather
than $[-5,5]$.
The weight, projection count and integration range follow
the minimal example in the LeJEPA repository
(\texttt{MINIMAL.md} at \url{https://github.com/galilai-group/lejepa}).

\paragraph{MAE comparison.} All ten runs use the same ViT-Nano, patch
size, optimizer schedule and batch size, with one render per example and
\eventMainHorizon{} epochs. MAE masks a fraction \methodMAEMask{} of the
patches and reconstructs only the masked ones, with mean squared error on
per-patch normalized targets. Its decoder has \methodMAEDepth{} blocks,
dimension \methodMAEDim{} and \methodHeads{} heads.

\paragraph{Readout implementation.} The readouts are fitted on \fitEvents{}
images and scored on \evalEvents{}. Log energy is standardized on the
fitting split, and the ridge standardizes all targets and features on that
split. The ridge regularization strength is \setupRidgeAlpha{}, fixed without
cross-validation. %
Fitting and evaluation images use
separate fixed seed bases, and training images are generated on the fly.
The canonical correlation analysis of Section~\ref{sec:direction} is
fitted by whitening and singular-value decomposition. %

The attention probe has about \methodQueryParams{} parameters. A single query,
which carries only the position of the image's fixed centre, attends to
the raw token grid. The probe has its own LayerNorm and a learnable
locality prior. It is trained with Adam at learning rate
$2\times10^{-3}$, batch size 128, for up to 2,500 steps. The MLP has
widths $96\to768\to768\to3$ with ReLU activations and is trained with
Adam at $10^{-3}$, batch size 256, for up to 4,000 steps. Both minimize
squared error on standardized targets, and we average two probe seeds.

We evaluate each probe every hundred steps and keep the fit with the
lowest mean per-target median absolute error on the evaluation split,
which also gives its reported $R^2$. An independent test split then
rescores these fits and pooled ridge, without reselecting. The token
checks fit a ridge at each token position, and a dual ridge to all tokens
concatenated (24,576 dimensions). The best position is chosen by mean
shape $R^2$ (the mean of the $z_1$ and $z_2$ $R^2$) on the evaluation split, which also gives its shape and energy
scores. The token readouts were not rescored on the independent test
split.

\paragraph{Direction calculation.}\label{app:direction}
We form second-moment matrices of the ridge residuals at the pre-onset and
final checkpoints, $\Sigma_e^{\rm pre}$ and $\Sigma_e^{\rm final}$, and of
the targets standardized on the fitting split, $\Sigma_y$, all on the
evaluation split without further centring. The leading generalized
eigenvector of $(\Sigma_e^{\rm final}-\Sigma_e^{\rm pre},\Sigma_y)$ is the
target combination whose residual error grew most relative to its second
moment. The reference direction $\wbarvec$ is the normalized mean of the
directions of the first four wide-projector runs, expressed in standardized
$z_1$, $z_2$ and $\log E$.

\paragraph{Onset rule.} We fixed the onset rule of
Section~\ref{sec:rig} after inspecting the readability trajectories of
our first 131 runs, and applied it unchanged to 16 runs trained later.
Over all 147 runs it marks 56 declines.

\paragraph{SIGReg estimator.} SIGReg averages the Epps--Pulley statistic
\citep{epps1983normality} over 256 random one-dimensional projections. The
statistic compares the empirical characteristic function of each
projection with that of a standard normal. Its integral uses 17-point
trapezoidal quadrature with a Gaussian window.

\section{Readout and Direction Checks}\label{app:eventchecks}

The independent test split contains \eventTestSize{} new images. Rescored
on it without reselection, pooled ridge and both learned probes changed by
less than \eventTestDifference{} in mean shape $R^2$. Both learned probes
still showed the decline, and each scored at or below pooled ridge at the
final checkpoint in \eventLearnedBelow{} runs, for shape and for log energy
separately.

Most wide-projector decline runs rebounded slightly from their trough (by at
most 0.085), and all ended below their earlier peak. Between the
pre-onset and final checkpoints, RankMe rose in all \eventAnatomySize{}
runs. After subtracting the feature mean, RankMe and its form on squared
singular values (Table~\ref{tab:metricconventions}) each rose in
\eventRankCentred{}, and the participation ratio
$(\sum_i s_i^2)^2/\sum_i s_i^4$ of the singular values $s_i$ in
\eventRankPR{}. RankMe on the projector output had
medians above 15.9, out of a maximum of \eventProjectorDim{}, at both
checkpoints, using BatchNorm statistics from the whole set
(Appendix~\ref{app:metricdetails}).

In the \eventAnatomySize{} runs, canonical correlation analysis between the
features and the three factors kept its first two correlations above
\eventLeadingCCA{} while the third fell, and the largest absolute eigenvalue of the
whitened change in residual covariance, from the pre-onset to the final
checkpoint, was \eventRankOneRange{} of the sum of absolute eigenvalues. In
\eventSweepAligned{} of the \eventSweepSize{} declines, the direction nearly
matches the reference direction (Appendix~\ref{app:direction}): the
smallest absolute cosine among them is \cosfloorRounded{}. The exception
declined twice. Split after inspection at epoch 300, its first decline has
cosine 0.997 and its second 0.11 or less.

A run's directional loss is the drop in directional
$R^2=1-w^\top\Sigma_e w/(w^\top\Sigma_y w)$ along its direction $w$, from
the pre-onset to the final checkpoint. The null comparison applied the same calculation to
\eventNullWindows{} windows from \eventNullRuns{} runs without a decline.
Each window ends at the run's final checkpoint and starts at the earliest,
median or latest pre-onset epoch of the declines with the same projector (for the
median, the checkpoint at or before it), when that epoch comes before the
end. Windows share runs, and all comparisons share the reference direction, so
these are not independent observations.

\paragraph{Cross-view signal-to-noise.} Unlike the other checks, this one
renders each probe image twice, as in a training pair. A ridge fitted on
one render predicts the standardized targets from both renders of each
evaluation image. For a
target direction $w$, write $a_i=w^\top\hat y_i^{(1)}$ and
$b_i=w^\top\hat y_i^{(2)}$. We used
$\mathrm{SNR}(w)=\operatorname{Var}_i[(a_i+b_i)/2]/
[\tfrac12\langle(a_i-b_i)^2\rangle_i]$ and compared the reference direction
with \methodRandomDirections{} random unit directions. We computed it for an
untrained encoder and for one default-projector run ten epochs before its onset. In
both, the reference direction was in the bottom tenth by SNR, and the
direction with the lowest SNR had absolute cosine \mechanismSNRInitCos{}
(untrained) and \mechanismSNRPreCos{} (pre-onset) with it. A separate comparison, on the untrained
encoder and on the oracle features of Appendix~\ref{app:occurrence}, used
directional $R^2$. It placed the
reference direction among the harder directions to read, but not the
hardest. In both, the hardest direction contrasts $z_1$ with $z_2$ and gives
$\log E$ an absolute weight of at most 0.17, against 0.567 in the reference
direction.

\paragraph{Pre-onset checkpoints.} In 21 of the 56 declines, the pre-onset
checkpoint comes before the peak, while readability is still rising, and
sits at most 28\% of the drop below the peak. In the other 35, it lies at
most 16\% of the way into the decline.

\section{Occurrence and Comparison Details}\label{app:occurrence}

\paragraph{Training views.} Partial shape refresh transforms the Gaussian
score $u$ behind each shape coordinate into $u'=\rho u+\sqrt{1-\rho^2}\eta$,
with independent standard-normal $\eta$, and then maps $u'$ to the chosen
shape distribution by matching quantiles. At $\ndial=0$ the scores are the
shape coordinates themselves. Energy stays shared, and $\rho=1$ gives shared views.

\begin{table}[!htbp]\centering\small
\caption{Training conditions not shown in Figure~\ref{fig:comparisons}, all
at $N=100$. Declines are runs whose drop from peak readability is at least
\readDeclineCut{}; drop ranges are rounded to two significant figures. LR is
the learning-rate schedule. MAE has no projector.}
\label{tab:occurrence-summary}
\begin{tabular}{@{}lrrrrl@{}}
\toprule Condition & Width & Epochs & Runs & Declines & Drop range \\ \midrule
$\ndial=0$, $\rho=0.9$ & 384 & 1000 & 6 & 0 & 0.00023--0.012 \\
$\ndial=0$, no weight decay & 384 & 300 & 3 & 1 & 0--0.29 \\
$\ndial=0$, constant LR & 384 & 300 & 3 & 2 & 0.027--0.27 \\
$\ndial=0$, SIGReg weight 0.05 & 384 & 300 & 10 & 3 & 0--0.27 \\
$\ndial=0.9$, cosine LR & 384 & 1000 & 3 & 3 & 0.21--0.55 \\
$\ndial=0.9$, constant LR & 384 & 1000 & 3 & 3 & 0.28--0.31 \\
$\ndial=0.97$, cosine LR & 384 & 1000 & 3 & 0 & 0.0019--0.036 \\
$\ndial=0.97$, $\rho=0.9$ & 384 & 300 & 3 & 0 & 0.0023--0.023 \\
ViT-Tiny, $\ndial=0$ & 2048 & 300 & 2 & 1 & 0--0.094 \\
MAE, $\ndial=0$ & -- & 300 & 10 & 0 & 0 \\
\bottomrule
\end{tabular}
\end{table}

Figure~\ref{fig:comparisons} leaves out six wide-projector runs trained
for 150 epochs: three at $\ndial=0$, which all declined, and
three at $\ndial=0.97$, which did not. No shared-view run at
$\ndial=0$ was trained for 1000 epochs, so the 1000-epoch refresh runs have
no matched comparison.

\paragraph{Element count.} An untrained encoder reaches readability
\untrainedThirty{}, \untrainedHundred{}, \untrainedThreeHundred{} and
\untrainedThousand{} at $N=30$, 100, 300 and 1000. At $N=300$ the median
peak sits \roomThreeHundred{} above that level, room for a typical drop. At
$N=1000$ it sits only \roomThousand{} above, less than a typical drop. At $N=30$ the three
declines dropped by \nThirtyDrops{}, against \nHundredDrops{} for the nine
declines at $N=100$. The other three runs peaked at \nThirtyLowPeaks{},
against \nThirtyHighPeaks{} for the three that declined.

\paragraph{Oracle readout.} The oracle reads the factors from physics
summaries of each image instead of encoder features. For each layer it uses
the total raw intensity, the intensity-weighted mean coordinates, their
variances and their covariance (\methodOracleFeatures{} features), computed
after undoing the input log transform. It uses the same ridge procedure as
the encoder readout.

\section{Feature Metrics and Checkpoint Selection}\label{app:metricdetails}

The seven metrics of \citet{arputharaj2026metrics} were computed on the
full fitting split, in \auditVariants{} variants
(Table~\ref{tab:metricconventions}). The metrics come from these papers:
$\alpha$-ReQ \citep{agrawal2022alphareq}, RankMe \citep{garrido2023rankme},
NE-Sum \citep{he2022exploring}, $\kappa$ and Self-Cluster
\citep{tsitsulin2023unsupervised}, DSE \citep{liao2024assessing} and TwoNN
\citep{facco2017estimating,ansuini2019intrinsic}. Checks against the original authors'
reference code passed. On synthetic power-law spectra, $\alpha$-ReQ read
0.009 to 0.059 high, most for the flattest spectrum; one of these checks,
0.559 for a true 0.5, exceeded its tolerance of 0.05. Figure~\ref{fig:response}'s
$\alpha$-ReQ medians lie between 4.7 and 6.4. $\kappa$ near the
singular-value cutoff and TwoNN in high dimensions are sensitive to how
they are estimated.

\begin{table}[!htbp]\centering\small
\caption{Metric variants, and the direction in which users read each
metric as better. Centred variants subtract the feature mean first.}
\label{tab:metricconventions}
\begin{tabular}{@{}p{.14\linewidth}p{.69\linewidth}p{.11\linewidth}@{}}
\toprule Metric & Variants & Preference \\ \midrule
$\alpha$-ReQ & Covariance-eigenvalue power-law fit over one-based
indices $10$ to $\lfloor0.9d\rfloor$: centred OLS, the original authors'
weighted fit, and uncentred OLS. Published code windows vary. & Closest to 1 \\
RankMe & Exponential of the entropy of normalized singular values, as
published or centred; also centred, on squared singular values. & Higher \\
NE-Sum & Sum of covariance eigenvalues divided by their maximum,
with and without centring. & Higher \\
$\kappa$ & Largest / smallest retained singular value, with and
without centring; cutoff uses ten times estimated float32 storage
rounding of the raw matrix. & Higher \\
Self-Cluster & Printed unsquared and normalization-consistent
squared forms, with and without centring. & Lower \\
DSE & Original authors' code and paper normalizations, matching either the
numerical bandwidth parameter or actual kernel width; row-normalized
variants. The original paper leaves bandwidth unspecified. & Higher \\
TwoNN & Line fit and maximum likelihood, averaged over twenty
90\% subsamples. & Lower \\
\bottomrule\end{tabular}
\end{table}

\paragraph{Projector path.} RankMe and LiDAR on the projector output apply
the saved projector to the mean encoder token, with BatchNorm statistics
from the whole probe set or the saved running statistics; neither
reproduces training, where BatchNorm saw paired minibatches.

\paragraph{Selection.} Candidates start at epoch 30, the first saved
checkpoint after epoch 20, and ties go to the earliest. Cost is computed for the average readability and for
each factor's $R^2$. For loss minimization in the decline runs, the median
cost was \costZone{} for $z_1$, \costZtwo{} for $z_2$ and \costE{} for
$\log E$; readability in these runs peaked at a median of \readPeakMedian{}
and ended at \readFinalMedian{}. In the runs without a decline, $\alpha$-ReQ
cost a median \selectionAlphaFlat{}, against \selectionFinalFlat{} for the final
checkpoint. The population is 28 decline runs and 22 LeJEPA runs
without a decline (the eight control runs of Appendix~\ref{app:chronological}, the alarm audit's 13 and
one ViT-Tiny run), with MAE kept separate. This population differs from
the alarm audit's (Appendix~\ref{app:chronological}). The loss, SIGReg
and feature standard deviation average the recorded iterations of each
epoch. The loss's fitted trend was decreasing in all 21 pre-onset windows
with at least four checkpoints (seven decline runs have fewer before their
onset) and in all 28 post-onset windows, though individual steps could
rise.

\paragraph{Figure 3.} Figure~\ref{fig:response} takes its runs from
the alarm audit (Appendix~\ref{app:chronological}), and preferred
directions follow Table~\ref{tab:metricconventions}. Of its
\respDecline{} decline runs, \respDeclineMuZero{} have $\mu=0$, against
\respHealthyMuZero{} of its \respHealthy{} runs without a decline. Each metric is computed as
\citet{arputharaj2026metrics} computed it, except Self-Cluster: its printed
form barely varies on these runs, so we use its squared form. NE-Sum is
centred and DSE uses the original authors' code normalization. Their other variants
give the same ranking after the onset. Three decline runs were trained for
only 150 epochs, so from 50 epochs after onset the decline curves use 22 to 24
runs. Aligning the runs without a decline anywhere from epoch 50 to 160 leaves
the verdicts of Section~\ref{sec:event} unchanged from 40 epochs after the
onset onward for six of the eight signals that rate the decline runs better.
SIGReg's verdict depends on the choice. Self-Cluster's changes at one
checkpoint under two of these alignments. The mixed loss and $\kappa$ also
depend on it.
Aligned instead at
each of the \respDecline{} onsets in turn, nine signals rate the decline runs
better after the onset: all except $\kappa$ and $\alpha$-ReQ. The loss does so
at all but one checkpoint, the other eight at every checkpoint. $\alpha$-ReQ
rates them worse.
Before the onset, the ranking depends on the alignment for every series.

\section{Alarm Audit}\label{app:chronological}

We audited the training
loss, SIGReg, feature standard deviation, LiDAR, the \auditVariants{}
variants of the seven metrics, RankMe and LiDAR on the projector output,
and LiDAR with groups that redraw shape (Section~\ref{sec:lidar}). On each series we ran three standard
change-point detection methods, CUSUM, EWMA and Page--Hinkley, with
thresholds set on separate control runs at $\ndial=0.97$. Alarms were
scored on 41 runs: the ten default runs, the wide-projector runs at otherwise
default settings (ten trained for 300 epochs and three for 150), and the
first three seeds per projector at $\ndial=0.4$, 0.7 and 0.9. Of these, 28
declined and the other \auditHeldout{} did not. Before any fitting,
wide-projector series are cut at epoch 150 and default-projector series at 300. Two
wide-projector decline runs have onsets at 150 and 160, so the window within ten
epochs of their onsets extends past the cut; the matched analyses use the
other \auditMatched{} decline runs. Each projector has four control runs at
$\ndial=0.97$. Thresholds came from two procedures: an initial one that
standardizes each run against its own early epochs, and a
control-referenced check that compares each run with the controls' median
change at the same epoch. We designed the second after seeing the first's
results, and neither has a nominal false-alarm guarantee.

\begin{figure}[!htbp]\centering
\includegraphics[width=\linewidth]{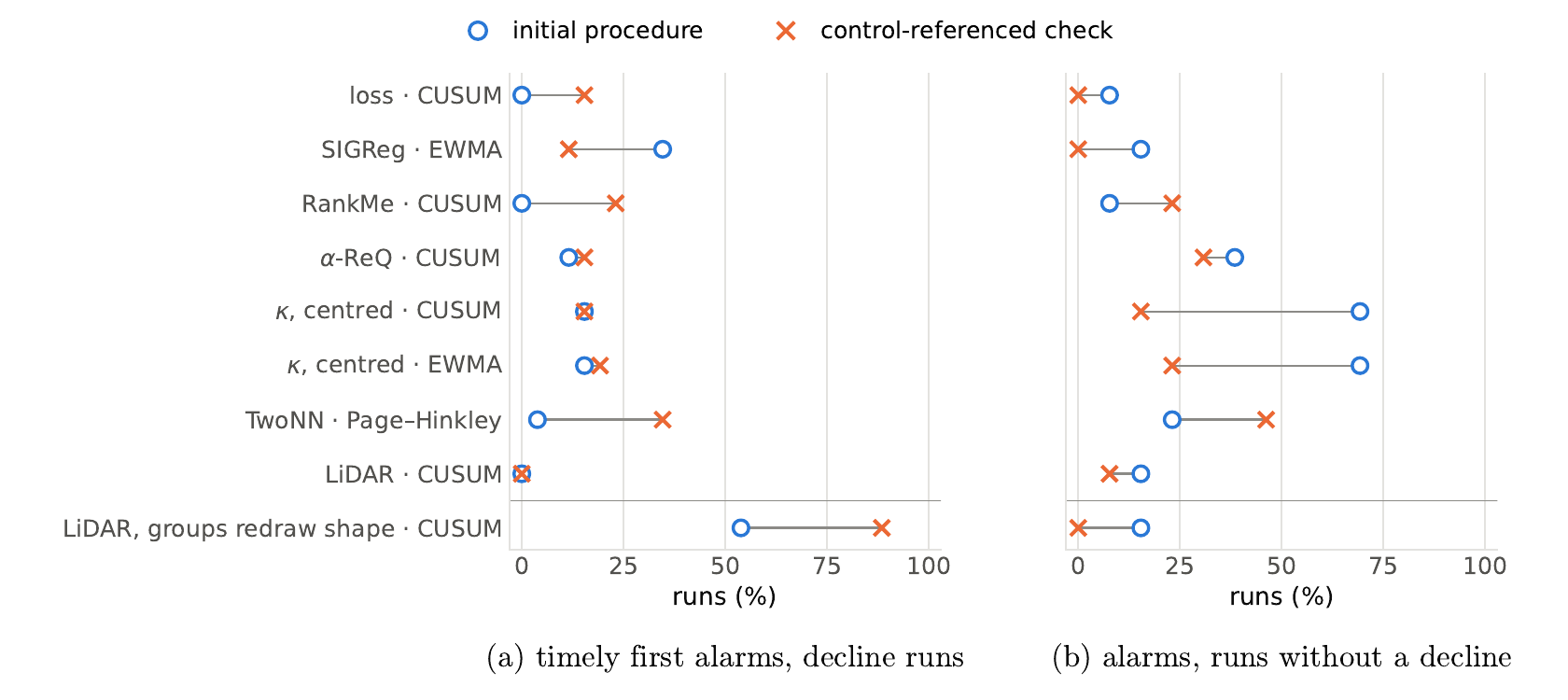}
\caption{Selected signal--test pairs at matched horizons: wide-projector
runs cut at 150 epochs and default-projector runs at 300, before fitting. Timely first
alarms among the \auditMatched{} decline runs whose onset window lies
inside these horizons, and alarms among the \auditHeldout{} runs
without a decline, for the initial and control-referenced procedures.
The last row uses groups that redraw shape (Section~\ref{sec:lidar}),
an augmentation not used in training. Centred $\kappa$ subtracts the
feature mean first.}
\label{fig:auditcomparison}
\end{figure}

The control reference helped centred $\kappa$, which raised far fewer
alarms in runs without a decline for about the same number of timely ones
(Figure~\ref{fig:auditcomparison}). With the control reference, TwoNN
under Page--Hinkley gained timely alarms, all in wide-projector runs, where
it also alarmed on most runs without a decline. Over variants, methods
and both procedures, the best timely count per metric was 9 of
\auditMatched{} for TwoNN and for $\alpha$-ReQ computed without
subtracting the feature mean, 6 for RankMe, 5 for
$\kappa$, 4 for NE-Sum, and 2 for DSE and Self-Cluster. At these counts,
TwoNN also alarmed in six or seven of the \auditHeldout{} runs without a
decline, depending on the estimator, and that $\alpha$-ReQ in five. SIGReg also
reached 9, with EWMA under the initial procedure, six of them with the
wide projector, and alarmed in two of the \auditHeldout{} runs without a decline. The loss and
feature standard deviation reached 4, and LiDAR 3. The
initial procedure calibrates on epochs 30--60, the same epochs where it
starts scoring, and \auditCalibOnsets{} of the \auditMatched{} onsets fall
in that window. The control-referenced check avoids this by comparing each
run with the completed control runs.

The initial procedure standardizes each run by its own mean and population
standard deviation over epochs 30--60. It uses the changes between
checkpoints for CUSUM and EWMA, and the values themselves for
Page--Hinkley. Scoring starts at epoch 30.

The control-referenced check subtracts the change typically seen in the
control runs, then expresses what remains relative to their usual
variability. For a control set $C$ with changes
$d_c(e)=x_c(e)-x_c(e-10)$, the reference $r_C$, residuals $u_c$ and
scale $s_C$ are
\begin{align*}
 r_C(e)&=\operatorname{median}_{c\in C}d_c(e),\\
 u_c(e)&=d_c(e)-r_{C\setminus\{c\}}(e),\\
 s_C&=1.4826\operatorname{median}_{c,e}
       |u_c(e)-\operatorname{median}_{c',e'}u_{c'}(e')|.
\end{align*}
Residuals are pooled with equal weight over controls and epochs. A monitored run is
scored against all four controls, $z(e)=[d(e)-r_C(e)]/s_C$, from
epoch 30. To set the threshold, each control is scored against the other
three, so neither its reference nor its scale uses it.

CUSUM and EWMA take $z$ as input. Page--Hinkley takes its running sum from
epoch 30, $L(e)=\sum_{t=30}^{e}z(t)$, which follows the level of the
series with the controls' typical changes removed. With input $v_t$, the
three rules are
\begin{align*}
 S_t^+&=\max(0,S_{t-1}^++v_t-0.5),&
 S_t^-&=\max(0,S_{t-1}^--v_t-0.5),\\
 C_t&=\max(S_t^+,S_t^-),&
 A_t&=0.2v_t+0.8A_{t-1},\\
 M_t&=M_{t-1}+v_t-0.5,&
 P_t&=\max(M_t-\min_{0\leq j\leq t}M_j,
                 \max_{0\leq j\leq t}M_j-M_t).
\end{align*}
All states start at zero. EWMA reports $|A_t|$ and Page--Hinkley reports
$P_t$. Because of its 0.5 slack, Page--Hinkley grows even for zero input,
so its threshold uses the same horizon as the monitored runs.

For each series, method and projector, the threshold is 1.1 times the largest
control statistic. The alarm is the first checkpoint at which the
statistic exceeds the threshold, and it is timely if it falls within ten
epochs of onset on either side.

\section{LiDAR Groups and Checks}\label{app:lidarchecks}

We compute LiDAR as defined by \citet[\S3]{thilak2024lidar}, on the
pooled encoder features of 64 groups of 8 renders at $N=100$. The
definition adds a small constant $\delta$ to the within-group covariance
without fixing its value, and we use $\delta=10^{-4}$. The LiDAR paper
recommends more groups than feature dimensions \citep[App.~11]{thilak2024lidar},
but we use 64 groups on \methodNanoDim{}-dimensional features, so at most 63
of LiDAR's eigenvalues are nonzero here. All runs, the
$\ndial=0.97$ controls included, use the same renders, with Gaussian
shapes, at every checkpoint. Section~\ref{sec:lidar} describes the two kinds
of groups.

In all 28 decline runs of the alarm audit (Appendix~\ref{app:chronological}),
the leading eigenvalue of LiDAR's spectrum, with groups that redraw shape,
fell from the pre-onset to the final checkpoint. Since $\log\mathrm{LiDAR}$
is the entropy of the eigenvalue shares, it splits as
$\log\mathrm{LiDAR}=h(q)+(1-q)H_{\rm rest}$, where $q$ is the leading
eigenvalue's share, $H_{\rm rest}$ the entropy of the other shares and $h$
the binary entropy. Changing $q$ from its pre-onset to its final
value, with $H_{\rm rest}$ held at either value and the two results
averaged, accounts for a median \lidarShareRounded{} of the change in log
LiDAR. Projections onto the leading direction
correlate with the ridge log-energy prediction at a median absolute
\lidarCorrRounded{}. With training-view groups the score also rose, in
\lidarSharedRise{} of these runs, but the leading eigenvalue fell in
only \lidarSharedLambdaFall{}.

Adding noise to the frozen features of the alarm audit's 13 runs without a
decline and the eight controls gives the same picture. With groups that
redraw shape, noise along the energy readout direction, at the size of the
median energy decline, produced timely initial-CUSUM alarms in
\lidarNoiseEnergy{} runs over three noise draws, and noise along the shape
readout direction produced none. The test perturbs features, not training.

Because $\delta$ is fixed, LiDAR depends on the scale of the features, so
we repeated the audit at $\delta=10^{-6}$ and $10^{-8}$. With the
control-referenced check and groups that redraw shape, at every $\delta$
CUSUM and EWMA on LiDAR flagged 23 to 25 of the \auditMatched{} scored
declines in time, with at most one false alarm among the \auditHeldout{}
runs without a decline; with training-view groups they flagged at most one
decline. The contrast between the two kinds of groups also held with a second,
independently drawn set of renders, and with renders whose shapes come
from each run's own training distribution.

Used as a quality score with groups that redraw shape, LiDAR chose a
checkpoint at or after onset in all 28 decline runs, at about the same
median readability cost as loss.

\end{document}